\documentclass[conference]{IEEEtran}

\newif\ifarxiv
\arxivtrue 

\usepackage{times}

\usepackage{microtype}
\usepackage{amsmath}
\usepackage{amssymb}
\usepackage{graphics}
\usepackage[tight]{subfigure}
\usepackage{tabularx}
\usepackage{booktabs}
\usepackage{multirow}
\usepackage{xspace} 
\usepackage[ancient]{flushend}
\usepackage[ruled]{algorithm2e}
\usepackage{bm}

\usepackage{acronym}
\acrodef{MDP}{Markov Decision Process}
\acrodef{POMDP}{Partially Observable Markov Decision Process}
\acrodef{DNN}{deep neural network}
\acrodef{CNN}{convolutional neural network}
\acrodef{RNN}{recursive neural network}
\acrodef{LSTM}{long short-term memory}
\acrodef{IRL}{inverse reinforcement learning}
\acrodef{IOC}{inverse optimal control}

\usepackage{color}
\usepackage[textsize=scriptsize]{todonotes} 
\newcommand{\lunarlander}[0]{{Lunar Lander}\xspace}
\newcommand{\lunarreacher}[0]{{Lunar Reacher}\xspace}
\newcommand{\dronereacher}[0]{{Drone Reacher}\xspace}
\newcolumntype{Y}{>{\centering\arraybackslash}X}

\usepackage{url}

\makeatletter
\let\NAT@parse\undefined
\makeatother

\usepackage[numbers,sort&compress]{natbib}
\bibpunct{[}{]}{,}{n}{}{;}

\usepackage{multicol}
\usepackage[bookmarks=true]{hyperref}

\ifarxiv
\fi

\begin{document}

\title{Residual Policy Learning for Shared Autonomy}

\ifarxiv
\author{\authorblockN{Charles Schaff}
\authorblockA{%
Toyota Technological Institute at Chicago\\
Chicago, IL 60637 USA\\
Email: cbschaff@ttic.edu}
\and
\authorblockN{Matthew R.\ Walter}
\authorblockA{%
Toyota Technological Institute at Chicago\\
Chicago, IL 60637 USA\\
Email: mwalter@ttic.edu}}
\else
\author{Author Names Omitted for Anonymous Review. Paper-ID 133}
\fi

\maketitle

\begin{abstract}

Shared autonomy provides an effective framework for human-robot collaboration that takes advantage of the complementary strengths of humans and robots to achieve common goals. Many existing approaches to shared autonomy make restrictive assumptions that the goal space, environment dynamics, or human policy are known a priori, or are limited to discrete action spaces, preventing those methods from scaling to complicated real world environments. We propose a model-free, residual policy learning algorithm for shared autonomy that alleviates the need for these assumptions. Our agents are trained to minimally adjust the human's actions such that a set of goal-agnostic constraints are satisfied. We test our method in two continuous control environments: \lunarlander, a 2D flight control domain, and a 6-DOF quadrotor reaching task. In experiments with human and surrogate pilots, our method significantly improves task performance without any knowledge of the human's goal beyond the constraints. These results highlight the ability of model-free deep reinforcement learning to realize assistive agents suited to continuous control settings with little knowledge of user intent.

\end{abstract}

\IEEEpeerreviewmaketitle

\section{Introduction}

Existing robot systems designed for unprepared environments generally provide one of two operating modes---full teleoperation (primarily in the field) or full autonomy (primarily in the lab). Teleoperation places significant cognitive load on the user. They must reason over both high- and low-level objectives and control the robot's low-level degrees-of-freedom using often unintuitive interfaces, while also interpreting the robot's various sensor streams. Tasks amenable to full autonomy are inhibited by a robot's limited proficiency at intervention (grasping and manipulation), long-term planning, and adapting to dynamic, cluttered environments. The ability to function in the continuum that exists between full teleoperation and full autonomy would enable operations that couple the complementary capabilities of humans and robots, improving the efficiency and effectiveness of human-robot collaboration.

Shared autonomy~\cite{abbink18} provides a framework for human-robot collaboration that takes advantage of the complementary strengths of humans and robots to achieve common goals. It may take the form of shared control, whereby the user and agent both control the same physical platform~\cite{goertz63, rosenberg93, Aigner97,dragan12,dragan13}, or human-robot teaming, whereby humans and robots operate independently~\cite{hoffman07,gombolay14} 
towards a shared goal. Early work in shared autonomy assumes that the user's goals are known to the agent, which is rarely realized in practice. Recent methods instead infer the user's goal from their actions and environment observations~\cite{muelling17, javdani15, perez15, hauser13, dragan13}. These methods often assume a priori knowledge of the environment dynamics and the set of possible goals, and require access to demonstrations or the user's policy for achieving each goal.

These assumptions can be limiting in practice, preventing the use of shared autonomy beyond simple tasks performed in structured, uncluttered environments. For example, estimating environment dynamics can be harder than learning to solve the task itself. Additionally, the goal space may be large, unknown, or may change over time. This will make it difficult or impossible to accurately infer the user's goal or learn the user's policy for each goal. At best, the tendency for goal inference to require the user to get close to the goal diminishes the advantages of shared autonomy. Inspired by the work of \citet{Reddy18} and \citet{broad17}, we seek to extend shared autonomy to more complicated domains through a framework in which the agent has no knowledge of the environment dynamics, the space of goals, or the user's intent.

\begin{figure}[!t]
    \centering
    \includegraphics[width=0.48\textwidth]{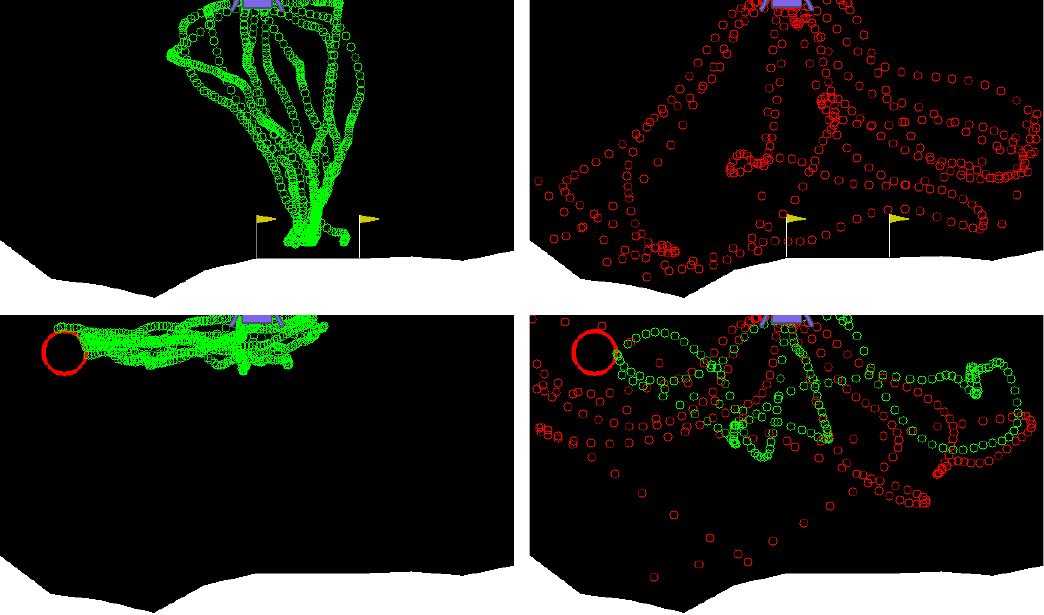}
    \caption{Human pilot trajectories in \lunarlander (top) and \lunarreacher (bottom) when controlling a spacecraft initialized with a random velocity with (left) and without (right) our shared autonomy agent. Trajectories rendered in green successfully landed between the flags or reached the target (large red circle). Red trajectories are those that crashed or went out of bounds. Circles are spaced evenly in time along each trajectory, with larger separation indicating greater velocity. The same assistant is used for both tasks and despite having no task-specific knowledge, it greatly improves the success rate of several humans and simulated pilots.}
    \label{fig:trajectories}
\end{figure}
To that end, we adopt a model-free deep reinforcement learning (RL) approach to shared autonomy.
Model-free deep RL has achieved great success on many complicated tasks such end-to-end sensor-based control~\cite{mnih13}, and robot manipulation and control~\cite{levine16, schulman17, levine18, tan18, duan16, schulman15a}.
We avoid assuming knowledge of human's reward function or the space of goals and instead focus on maintaining a set of goal-agnostic constraints.
For example, a human driver is expected to follow traffic laws and not collide with other vehicles, pedestrians, or objects regardless of the destination or task.
This idea is naturally captured by having the agent act to satisfy some criteria or set of constraints relevant to multiple tasks within the environment.
Without knowing the task at hand, the robot should attempt to minimally intervene while maintaining these constraints.

Shared autonomy methods differ in the manner in which the agent augments the control of the user, which requires balancing the advantages of increased levels of agent autonomy with a human's desire to maintain control authority~\cite{kim11}. This complicates the use of standard deep reinforcement learning approaches, which traditionally assume full autonomy. In an effort to satisfy the user's control preference, we approach shared autonomy from the perspective of residual policy learning~\cite{Silver2018ResidualPL, Johannink19}, which learns residual (corrective) actions that adapt a nominal ``hard-coded'' policy. In our case, the (unknown) human policy plays the role of the nominal policy that the residual shared autonomy agent corrects to improve performance. We find this to be a natural way to combine continuous control inputs from a human and an agent.

Using this method, we are able to create robotic assistants that improve task performance across a variety of human and simulated actors, while also maintaining a set of safety constraints. Specifically, we apply our method in two assistive control environments: Lunar Lander and a 6-DOF quadrotor reaching task. For each task, we conduct experiments with human operators as well as  with surrogate pilots that are handicapped in ways representative of human control. Trained only to satisfy a constraint on not crashing, we find that our method drastically improves task performance and reduces the number of catastrophic failures. Videos and code can be found at \href{https://ttic.uchicago.edu/~cbschaff/rsa/}{https://ttic.uchicago.edu/$\sim$cbschaff/rsa/}.

\section{Related Work}

There is a long history of work in shared autonomy across a variety of domains including remote telepresence~\cite{goertz63, rosenberg93, debus00}, assistive robotic manipulation~\cite{kim06, schroer15, muelling17},
and  assistive
navigation~\cite{argall16, ghorbel17}.
Early work assumes that the user's goal is
known~\cite{rosenberg93, crandall02, kofman05, you12} to the agent. Recent methods relax this assumption and instead treat the user's goal as a latent random variable, for example, by modeling shared autonomy as a relaxation of a partially observable Markov decision process (POMDP)~\cite{javdani15}. These methods~\cite{li03, yu05, kragic05, kofman05, aarno05, dragan13, hauser13, javdani15, perez15, muelling17, Broad19} 
exploit the user's actions and environment observations to infer the user's goal. Common approaches to predicting the user's goal formulate the problem in the context of Bayesian inference~\cite{li03, aarno05, kragic05, hauser13}, while many others use inverse reinforcement learning~\cite{abbeel04, ratliff06, ziebart08}.
These approaches typically assume knowledge of the environment dynamics, the space of goals, or the user's goal-dependent policy. However, these assumptions are often invalid in practice. One may only have noisy observations of the state (e.g., images and language), and the dynamics are typically unknown and may be more difficult to estimate than the target policy itself. Further, th space of goals may be unstructured.

Our work is inspired by recent work by \citet{Reddy18} and \citet{broad17}. \citet{Reddy18} attempts to remove some of these assumptions by using model-free deep reinforcement learning. Specifically, they do not assume knowledge of transition dynamics or the space of user goals. Instead, they  assume that the agent has access to a reward function relevant for all tasks in the environment and that the user provides task-specific feedback about the agent's performance at the end of each episode. They use this feedback to learn the optimal Q-function for each task. The agent takes the action that is closest to the human's action while being within a specified distance from the optimal action. However, the sample complexity of model-free deep reinforcement learning techniques is quite poor and it may be taxing or infeasible to require user feedback after each episode. Additionally, the assistant is trained separately for each task. Providing assistance for another task requires addition training. Additionally, the method is restricted to domains with discrete action spaces.

\citet{broad17} take a different approach to shared autonomy, using outer-loop stabilization to share control. They compare human actions to an optimal controller and only execute the human action when it is sufficiently close to the optimal control for the task. They do not assume knowledge of the environment dynamics and instead learn a dynamics model directly from human data. They use this model in a model predictive control framework to derive their optimal controller. This approach is applicable to continuous action spaces, but they assume full knowledge of the user's goal (i.e., the reward function) in order to derive their optimal controller.

Within the realm of shared autonomy, methods differ in how the agent's control authority interacts with the user's inputs. One approach is to explicitly switch from user control to full autonomy when certain conditions occur (e.g., when the goal prediction likelihood exceeds a threshold)~\cite{kofman05}.
In an effort to provide the user with more control, other methods seek to minimally augment user control to satisfy particular conditions (e.g., collision-avoidance)~\cite{crandall02, kim06, ghorbel17, Broad19} or to obey constraints~\cite{aarno05, kragic05, Anderson12, Erlien16, Schwarting17, Amini18}. Along these lines, policy blending~\cite{dragan13} uses an arbitration function to combine the user's input with the agent's predictions. We similarly seek to maintain the user's control authority by treating the agent's actions as minimal corrective actions necessary to satisfy constraints that generalize across tasks. However, we do not assume knowledge of the environment dynamics or the space of goals.

Our work builds off of a recent advance in reinforcement learning for robotics: residual policy learning~\cite{Johannink19, Silver2018ResidualPL}. Residual policy learning improves upon an initial policy by learning residual actions that are added to the actions of the initial policy. This can greatly reduce sample complexity in tasks where a decent but not great controller can be easily obtained. Our approach, however, leverages this as a way to share control between two agents. Our robotic assistant learns a residual action that is added to the action of the human. We can then constrain the assistant by regularizing the size of the residual.

\section{Background}

Before introducing our method, we first present relevant background material in reinforcement learning, shared autonomy, residual policy learning, and constrained MDPs.

\subsection{Reinforcement Learning}

In reinforcement learning~\cite{sutton18}, an agent interacts with a \acf{MDP} defined by the tuple $\mathcal{M}=\{\mathcal{S}, \mathcal{A}, \mathcal{T}, \mathcal{R}, \gamma\}$, with state space $\mathcal{S}$, action space $\mathcal{A}$, transition function $\mathcal{T}: \mathcal{S} \times \mathcal{A} \rightarrow \Delta(\mathcal{S})$ that maps state-action pairs to a distribution over next states, reward function $\mathcal{R}: \mathcal{S} \times \mathcal{A} \rightarrow \mathbb{R}$, and discount factor $\gamma \in [0, 1)$.

At each timestep, the agent observes the state $s \in \mathcal{S}$ and selects an action $a \in \mathcal{A}$. The environment then transitions to a new state $s' \sim \mathcal{T}(s, a)$ and the agent receives a reward $R(s, a)$. The goal of the agent is to find a policy \mbox{$\pi: \mathcal{S} \rightarrow \Delta(\mathcal{A})$} that maximizes the discounted sum of rewards, $J(\pi) = \mathbb{E}_\pi[\sum_{t=0}^T \gamma^t \mathcal{R}(s_t, a_t)]$. We can write this objective as
\begin{equation} \label{eqn:objective}
  \pi^* = \underset{\pi}{\text{arg max}} \; \mathbb{E}_\pi\left[\sum_{t=0}^T \gamma^t \mathcal{R}(s_t, a_t) \right].
\end{equation}

In this work, we use policy gradient-based methods~\cite{sutton00} to find an optimal policy. We assume that our policy is parameterized by a vector $\theta$, and compute the gradient of the objective $J(\pi_\theta)$ w.r.t.\ $\theta$ using the log-likelihood trick:
\begin{equation} \label{eqn:gradient}
\nabla_\theta J(\pi_\theta) = \mathbb{E}_\pi \left[\nabla_\theta \log \: \pi_\theta(s_t) \sum_{t=0}^T \gamma^t \mathcal{R}(s_t, a_t) \right].
\end{equation}
We can then optimize $\theta$ and, in turn, the policy by acting under $\pi_\theta$ and using stochastic gradient ascent. In this work, we build upon a slightly more sophisticated version of policy gradients called Proximal Policy Optimization (PPO)~\cite{schulman17}, which provides improved efficiency and robustness.

\subsection{Shared Autonomy}

In shared autonomy, a human and an agent share control of a system to achieve a
common goal $g \in \mathcal{G}$ from the space of possible goals $\mathcal{G}$. When the goal is
known to the agent, the problem can be formulated as an MDP $\mathcal{M} =
\{\mathcal{S} \times \mathcal{G} \times \mathcal{A}_h, \mathcal{A}_r,
\mathcal{T}, \mathcal{R}, \gamma\}$ in which the goal and the human action are part of the state space. Here,
 we differentiate between the action spaces of the human $\mathcal{A}_h$ and
that of the robot $\mathcal{A}_r$. In practice, however, the goal is often not
known to the agent. Existing methods mitigate this by treating the user's goal
as a latent random variable that is then inferred from observations~\cite{li03,
yu05, kragic05, kofman05, aarno05, koppula13, dragan13, hauser13,
javdani15, perez15, muelling17}.

In particular, the problem can be formulated as
a partially observable Markov decision process (POMDP)~\cite{javdani15}
$\mathcal{M}_r = \{\mathcal{S} \times \mathcal{G} \times \mathcal{A}_h,
\mathcal{A}_r, \mathcal{T}, \mathcal{R}, \Omega, \mathcal{O}, \gamma\}$, where
$\Omega = \mathcal{S} \times \mathcal{A}_h$ is the observation space and $\mathcal{O}: \mathcal{S} \times \mathcal{G} \times \mathcal{A}_h \rightarrow \Delta(\Omega)$ is the observation function. In this case, the observation function just returns the state without goal information. Existing approaches exploit environment observations to infer the user's goal using inverse reinforcement learning~\cite{ng00, abbeel04, ziebart08}
or hindsight optimization~\cite{javdani15}, among others.
This often requires the assumption that the goal space $\mathcal{G}$, transition function $\mathcal{T}$, or human policy $\pi_h$ are known. We do not make such assumptions.

\subsection{Residual Policy Learning}

Residual policy learning~\cite{Silver2018ResidualPL, Johannink19} attempts to use a baseline policy $\pi_0$ to reduce the sample complexity of reinforcement learning methods. The learned policy acts by adding a ``residual'' (corrective) action $a_r \sim \pi(s, a_0)$ to the action $a_0 \sim \pi_0(s)$ provided by the nominal policy. Formally, the residual agent acts in an MDP with state space and transition dynamics augmented by $\pi_0$: $\mathcal{M} = (\mathcal{S} \times \mathcal{A}, \mathcal{A}, \mathcal{T}', \mathcal{R}, \gamma)$, where $\mathcal{T}'([s,a_0], a_r, [s', a_0']) = \mathcal{T}(s, a_0 + a_r, s') \mathbb{P}(a_0' | \pi_0(s'))$.

The residual agent can be trained using any reinforcement learning algorithm that allows for continuous action spaces. One can think of residual policy learning as biasing exploration towards the state distribution of $\pi_0$. A reduction in the sample complexity required for training can be explained by a reduction in the exploration required to obtain high reward. In our work, $\pi_0$ is replaced with a human actor.

\subsection{Constrained MDPs}

A constrained Markov decision process (CMDP)~\cite{Altman99} is a multi-objective MDP with a set of constraints that the policy must satisfy. Let $\mathcal{C} = \{C_1, \ldots, C_m\}$ be a set of $m$ cost functions of the form: $C_i: \mathcal{S} \times \mathcal{A} \rightarrow \mathbb{R}$. Let $J_{C_i}(\pi) = \mathbb{E}_\pi[\sum_t \gamma^t C_i(s_t, a_t)]$ be the expected discounted cost under policy $\pi$ with respect to $C_i$. Let $d_i \in \mathbb{R}$ be a constant threshold associated with cost $C_i$. One can then define $\Pi_C$ to be the set of policies that satisfy all constraints:
\begin{equation}
    \Pi_C = \left \{ \pi : \forall i \in \{1,\ldots m\}, J_{C_i}(\pi) \leq d_i \right \}.
\end{equation}
In constrained MDPs, the goal is to find a policy $\pi \in \Pi_C$ that maximizes the discounted sum of rewards
\begin{equation}
  \pi^* = \underset{\pi \in \Pi_C}{\text{arg max}} \; \mathbb{E}_\pi\left[\sum_{t=0}^T \gamma^t \mathcal{R}(s_t, a_t) \right],
\end{equation}
while satisfying the constraints.

\section{Method}
\begin{figure}[!t]
    \centering
    \includegraphics[width=0.45\textwidth]{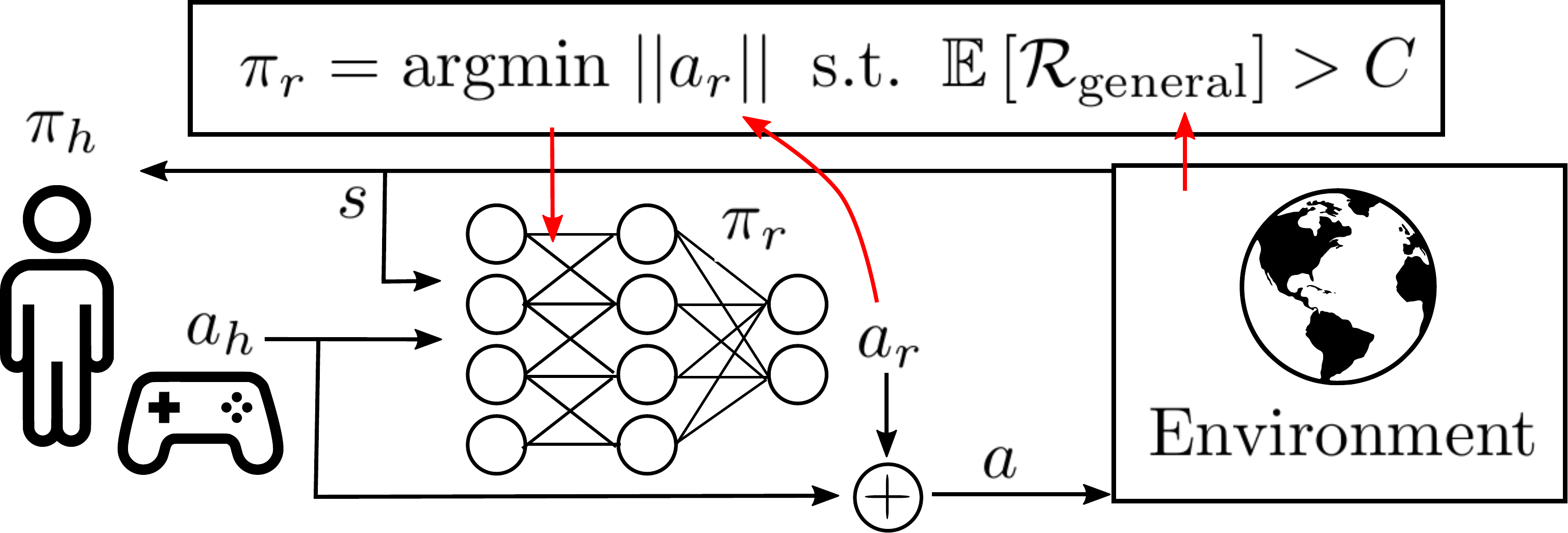}
    \caption{An overview of our residual policy learning algorithm for shared autonomy. A human takes an action $a_h \sim \pi_h(s)$ based on the current state $s$ according to an unknown policy $\pi_h(s)$. The agent then draws an action $a_r \sim \pi_r(s, a_h)$ from a learned policy according to the current state and user action. These actions are added together and executed in the environment, which produces the next state and a reward $R_{\text{general}}$, which captures objectives common to multiple tasks. The goal of the agent is to minimally adjust the human's actions such that the expected reward is above a given threshold.}
    \label{fig:method}
\end{figure}
In this section, we describe our method for residual shared autonomy. An overview of our method can be found in Figure~\ref{fig:method}.

\subsection{Problem Statement}

Recall from the previous section that we are operating in an MDP with the following form: $\mathcal{M} = (\mathcal{S} \times \mathcal{G} \times \mathcal{A}_h, \mathcal{A}_r, \mathcal{T}, \mathcal{R}, \gamma)$. When goal information is unavailable, we can think of this as a POMDP in which the current goal is hidden from the agent. Shared autonomy requires some mechanism that integrates the agent's control authority with that of the human. In this work, we formulate this arbitration in the context of residual policy learning~\cite{Silver2018ResidualPL, Johannink19}, whereby the agent's control input takes the form of a residual correction to the human's actions, \mbox{$a = a_h + a_r$},
where $a_h \sim \pi_h(s, g)$ and $a_r \sim \pi_r(s, a_h)$ are actions drawn from the human and agent policies, respectively. This formulation is naturally applicable in domains with continuous action spaces and offers the additional advantage that regularizing the residual encourages the agent to endow more control authority to the human whenever possible.

In order to consider domains with environment dynamics $\mathcal{T}$ that are unknown and possibly difficult to learn, we take a model-free approach to shared autonomy. To enable operation without any knowledge of the goal space $\mathcal{G}$, we assume access to a set of \emph{goal-agnostic} constraints, $\mathcal{C} = \{C_1, \ldots, C_m\}$, that should be satisfied. In the case of driving, example constraints would encourage avoiding collisions, staying in the lane, etc. Enforcing the constraints to be goal-agnostic removes the need to infer the human's goal, transforming the problem from a POMDP to a CMDP.

However, there may be many ways to satisfy the given constraints, and not all of them are desirable. For example, a trivial solution to satisfying the constraint associated with avoiding collisions is to cancel the human's control to keep the vehicle stationary. We address this issue by encouraging solutions that minimize the residual (corrective) actions taken by the agent, while satisfying the constraints.
Intuitively, this approach allows the agent to assist a human in whatever goal they decide to solve, while allowing the human to be responsible for all goal-directed decisions (e.g., where the car goes and what route takes). This includes situations in which the human changes goals arbitrarily or makes up entirely new goals (e.g., pulling into a parking space).

Formally, given an unknown human policy $\pi_h$, a set of constraints  $\mathcal{C} = \{C_1, \ldots, C_m\}$, and a set of corresponding thresholds $\mathcal{D} = \{d_1, \ldots, d_m\}$, we learn the agent's policy $\pi_r(s,a_h)$ by optimizing the following objective:
\begin{equation}
    \begin{split}
        \pi_r^* = \; &\underset{\pi_r}{\text{arg min}} \; \mathbb{E}_{\pi_r, \pi_h}\left[\lVert a_r \rVert \right] \\
        &\text{s.t. } \; \forall_i \; \mathbb{E}_{\pi_r, \pi_h}\left[C_i(s, a_h + a_r)\right] > d_i
    \end{split}
\end{equation}

While our formulation is applicable to arbitrary constraints, we focus on the setting in which all constraints can be expressed in a single goal-agnostic reward function: $\mathcal{R}_\textrm{general}$ \footnote{While we focus on "goal-agnostic" constraints, it is possible to include goal-specific information if desired.}. We assume access to $\mathcal{R}_\textrm{general}$ and maintain a single constraint on the return:
\begin{equation}
    \begin{split}
        \pi_r^* = \; &\underset{\pi_r}{\text{arg min}} \; \mathbb{E}_{\pi_r, \pi_h}\left[\lVert a_r \rVert \right] \\
        &\text{s.t. } \; J(\pi_r) > d
    \end{split}
\end{equation}
where $J(\pi_r) = \mathbb{E}_{\pi_r, \pi_h}\left[\sum_t \gamma^t \mathcal{R}_{\textrm{general}}(s_t, a^t_h + a^t_r)\right]$.
Writing out the Lagrangian of the above optimization problem, we obtain the following objective:
\begin{equation}
  \pi_r^* = \; \underset{\pi_r}{\text{arg min}} \;\; \underset{\lambda \geq 0}{\text{max}} \; \left\{\mathbb{E}_{\pi_r, \pi_h}\left[\lVert a_r \rVert \right]
  + \lambda \left ( d - J(\pi_r) \right )\right\}
\end{equation}
where $\lambda$ is the Lagrangian multiplier.

\subsection{Constrained Residual PPO for Shared Control}

\begin{algorithm}[t] \label{alg1}
\SetAlgoLined
 Initialize $\pi_r^\theta$, $\lambda$

 Initialize learning rates $\alpha_\theta$ and $\alpha_\lambda$

 Initialize rollout length $T$.

 \While{not converged}{
  \For{$t \in \{1, \ldots, T\}$}{

    Sample $a_h^t \sim \hat{\pi}_h(s_t)$.

    Sample $a_r^t \sim \pi_r^\theta(s_t, a_h^t)$.

    Execute action $a_t = a_h^t + a_r^t$ in environment.

    Observe $s_{t+1}$ and $r_t$.
  }
  Estimate losses $L_\theta(\theta, \lambda)$ and $L_\lambda(\theta, \lambda)$ using $\{s_1, \ldots, s_T\}$, $\{a_r^1, \ldots, a_r^T\}$, and $\{r_1, \ldots, r_T\}$.

  $\theta \leftarrow \theta - \alpha_\theta \nabla_\theta L_\theta(\theta, \lambda)$

  $\lambda \leftarrow \lambda + \alpha_\lambda \nabla_\lambda L_\lambda(\theta, \lambda)$
 }
 \caption{Constrained Residual Shared Autonomy}
\end{algorithm}

We simultaneously solve for $\lambda^*$ and $\pi_r^*$ using a constrained version of Proximal Policy Optimization (PPO)~\cite{schulman17}. Specifically, we parameterize $\pi_r$ in terms of a vector $\theta$, denoted as $\pi_r^\theta$. We then optimize over these parameters $\theta$ by minimizing the following objective with stochastic gradient descent:
\begin{equation}
    L_\theta(\theta, \lambda) = \frac{1}{1 + \lambda} \mathbb{E}_{\pi_r^\theta, \pi_h}\left[\lVert a_r \rVert \right]
    + \frac{\lambda}{1 + \lambda} L_\text{PPO}(\pi_r^\theta),
\end{equation}
 where $L_\text{PPO}$ is the surrogate object used in the PPO algorithm. We simultaneously optimize $\lambda$ by maximizing the following objective with stochastic gradient ascent:
 \begin{equation}
     L_\lambda(\theta, \lambda) = \text{softplus}(\lambda) (d - J(\pi_r^\theta)),
 \end{equation}
 where $\text{softplus}(x) = \text{log}(1 + e^x)$ constrains $\lambda$ to be positive. See Algorithm~\ref{alg1} for details.

\begin{figure*}[!t]
  \centering
    \includegraphics[width=0.95\textwidth]{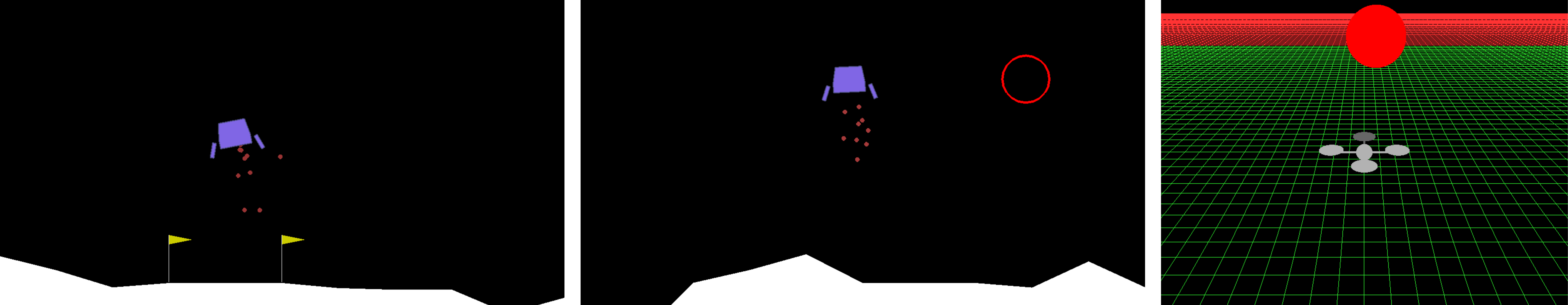}
    \caption{
    A visualization of the three continuous control environments that we use for evaluation. In \lunarlander (left), the goal is to land between the two flags. The goal in \lunarreacher (center) is to fly to the red circle. The goal in \dronereacher (right) is to fly to the red sphere.}
    \label{fig:envs}
\end{figure*}

\subsection{Learning with a Human}

Up until this point, we have ignored the effects of $\pi_h$ on our algorithm. While it is possible to train with a human in the loop, the sample complexity of model-free reinforcement learning can be prohibitive, particularly with on-policy methods such as PPO. In order to reduce the cost of and speed up training, we train the shared autonomy agent using a surrogate policy $\hat{\pi}_h$ in place of the human policy. This has been done with other model-free shared autonomy methods~\cite{Reddy18}. There are many ways to formulate this surrogate policy, which we believe should exhibit the following characteristics. First, when possible, we would like the agent to be able to assist several humans without the need to specialize or fine-tune to each individual. Therefore, we want a surrogate policy that exhibits a diverse set of human-like strategies. Second, as mentioned earlier, we would like the agent to generalize across goals. Therefore, the surrogate policy should encourage the agent to explore the state space as much as possible.

Following these desiderata, we design a surrogate human policy by imitating a set of human actors. Specifically, we asked several people ($9$ for \lunarlander and $14$ for \dronereacher) to perform the task and then trained a surrogate agent for each participant using behavioral cloning~\cite{bain95}. We then used all of the models to train our agent, switching between them periodically during training. Despite the fact that the learned models performed very poorly on the given tasks, we found this approach to greatly improve the robustness of the agent over other surrogates, such as using an optimal policy or an optimal policy that has been handicapped to be more ``human-like'' (i.e., by adding noise or by adding latency by forcing it to repeat actions), as has been done elsewhere~\cite{Reddy18}.

\section{Results}

\begin{table*}[!t]
    \centering
    \begin{tabularx}{0.3\linewidth}{lYYY}
        \multicolumn{4}{c}{Success Rate}\\
        \toprule
        Copilot & Laggy Pilot & Noisy Pilot & Imitation Pilot \\
        \midrule
        None & $0.389$ & $0.199$ & $0.003$\\
        Laggy & $0.552$ & $0.493$ & $0.009$\\
        Noisy & $\bm{0.828}$ & $\bm{0.866}$ & $\bm{0.147}$\\
        Imitation & $0.624$ & $0.602$ & $0.003$\\
        \bottomrule
    \end{tabularx}\hfil
    \begin{tabularx}{0.3\linewidth}{lYYY}
        \multicolumn{4}{c}{Crash Rate}\\
        \toprule
        Copilot & Laggy Pilot & Noisy Pilot & Imitation Pilot \\
        \midrule
        None & $0.567$ & $0.794$ & $0.996$\\
        Laggy & $0.397$ & $0.495$ & $0.988$\\
        Noisy & $\bm{0.073}$ & $\bm{0.036}$ & $0.481$\\
        Imitation & $0.293$ & $0.161$ & $\bm{0.015}$\\
        \bottomrule
    \end{tabularx}\hfil
    \begin{tabularx}{0.3\linewidth}{lYYY}
        \multicolumn{4}{c}{Reward}\\
        \toprule
        Copilot & Laggy Pilot & Noisy Pilot & Imitation Pilot \\
        \midrule
        None & $\hphantom{0}86$ & $\hphantom{0}74$ & $-145$\\
        Laggy & $154$ & $147$ & $-104$\\
        Noisy & $\bm{242}$ & $\bm{250}$ & $\bm{-85}$\\
        Imitation & $186$ & $170$ & $-101$\\
        \bottomrule
    \end{tabularx}
    \caption{\lunarlander performance}
    \label{tab:lunar_sim}
\end{table*}
\begin{table*}[!t]
    \centering
    \begin{tabularx}{0.3\linewidth}{lYYY}
        \multicolumn{4}{c}{Success Rate}\\
        \toprule
        Copilot & Laggy Pilot & Noisy Pilot & Imitation Pilot \\
        \midrule
        None & $0.676$ & $0.772$ & $0.018$\\
        Laggy & $0.060$ & $0.008$ & $0.005$\\
        Noisy & $0.700$ & $\bm{0.917}$ & $0.004$\\
        Imitation & $0.686$ & $0.868$ & $0.003$\\
        \bottomrule
    \end{tabularx}\hfil
    \begin{tabularx}{0.3\linewidth}{lYYY}
        \multicolumn{4}{c}{Crash Rate}\\
        \toprule
        Copilot & Laggy Pilot & Noisy Pilot & Imitation Pilot \\
        \midrule
        None & $0.234$ & $0.212$ & $0.900$\\
        Laggy & $\bm{0.008}$ & $0.075$ & $0.183$\\
        Noisy & $0.202$ & $\bm{0.002}$ & $0.758$\\
        Imitation & $0.200$ & $\bm{0.002}$ & $\bm{0.001}$\\
        \bottomrule
    \end{tabularx}\hfil
    \begin{tabularx}{0.3\linewidth}{lYYY}
        \multicolumn{4}{c}{Reward}\\
        \toprule
        Copilot & Laggy Pilot & Noisy Pilot & Imitation Pilot \\
        \midrule
        None & $-24$ & $-25$ & $-109$\\
        Laggy & $\bm{-0.3}$ & $-7.7$ & $-19$\\
        Noisy & $-21$ & $\hphantom{0}\bm{5.5}$ & $-82$\\
        Imitation & $-21$ & $\hphantom{0}\bm{5.9}$ & $\bm{-0.1}$\\
        \bottomrule
    \end{tabularx}
    \caption{Drone Performance}
    \label{tab:drone_sim}
\end{table*}

We evaluate the effectiveness of our shared autonomy algorithm through three control tasks (Fig.~\ref{fig:envs}):

\textsc{Lunar Lander}: A 2D continuous control environment from OpenAI Gym~\cite{brockman16}. The goal of the game is to land a small spaceship at a specified location on an uneven ground without crashing. The terrain, goal location, and initial velocity vary randomly between episodes. Each episode ends when the spaceship lands and becomes idle, crashes, flies out of bounds, or a timeout of $1000$ steps is reached. The state space is $8$ dimensional, consisting of the position, velocity, angular position, angular velocity, and whether or not each leg is in contact with the ground. The two-dimensional action space consists of vertical and lateral thrusts, both of which are continuous. $\mathcal{R}_{\text{general}}$ penalizes crashing, going out of bounds, and has a small penalty on fuel usage. A shaping term encourages exploration of policies that have low velocity and stay upright. $\mathcal{R}_{\text{g}}$ rewards landing at the goal. For all three tasks, $\mathcal{R}_{\text{g}}$ is only used to train the surrogate human policies.

\textsc{Lunar Reacher}: A modified version of \lunarlander. The state and action spaces as well as the dynamics remain the same, but the goal is now to fly to a specified target. We use this task to test the generalization of our trained agents across a wider range of goals. $\mathcal{R}_{\text{general}}$ is the same as \lunarlander and $\mathcal{R}_{\text{g}}$ rewards the agent for reaching the target. A similar environment was designed and used by \citet{broad17}.

\textsc{Drone Reacher}: A simulated environment in which a human must fly a 6-DOF quadrotor to a random floating target. The state space is $15$-dimensional, encoding the drone's position, orientation, linear velocity, and angular velocity, and the goal position. The action space is four-dimensional and continuous, including thrust along the $z$-axis of the drone and torque along the $x$-, $y$-, and $z$-axes of the drone. The simulator applies a small amount of random force and torque to the drone to emulate external disturbances (e.g., wind gusts). Each episode ends when the target is reached, the drone crashes or flies out of bounds, or a timeout is reached.  $\mathcal{R}_{\text{general}}$ penalizes crashing and going out of bounds. It also has a shaping term to encourage exploration of policies that have low velocity and achieve level flight (pitch and roll near zero). $\mathcal{R}_{\text{g}}$ rewards the agent for reaching the target.

In all three domains, we refer to the human operator as the \textit{pilot} and the agent as the \textit{copilot}. We evaluate the effectiveness of our copilots both quantitatively and qualitatively. To do so, we compare the performance of different simulated pilots when operating with and without the assistance of a copilot, and by conducting experiments in which human subjects are asked to complete the various tasks with and without assistance.

We emphasize that in every domain, we train the copilot with no knowledge of the specific task objectives and that the same copilot is used for all pilots. We do not fine-tune or otherwise specialize the copilot for a given pilot or task.

\subsection{Experimental Details}

We first collect user demonstrations for each environment to train our surrogate human policies. For \lunarlander, we collected $100$ episodes from $9$ participants. All participants were allowed to familiarize themselves with the environment for as long as they wanted before data was collected. For \dronereacher, we collected $30$ episodes from $14$ participants in the same manner. We then train a three-layer neural network with $128$ hidden units per layer using behavioral cloning to imitate each participant. Importantly, participants were novices and the imitation agents learned to make ``human-like'' errors, which is valuable when training the copilot. While training the copilot, our surrogate human policy $\hat{\pi}_h$ randomly switched between imitation agents at each timestep with probability $0.001$.

We model the copilot as a three-layer neural network with $128$ hidden units per layer. The network has two heads, one for the policy (outputting the mean and variance of a Gaussian distribution) and one for the value function. We trained the network according to the procedure outlined in Algorithm~\ref{alg1} for $100$\,M timesteps. As with other residual policy learning approaches, we use a warm-up period of $100$\,K timesteps in which the copilot outputs zeros and the network is trained to estimate the value function of the surrogate pilot. We decayed the learning rates $\alpha_\theta$ and $\alpha_\lambda$ every $20$\,M timesteps by a factor of $\sqrt{.1}$.

To generate simulated pilots, we trained an RL agent for each task using PPO~\cite{schulman17} for \dronereacher and TD3~\cite{fujimoto18} for \lunarlander.
We then handicap the trained pilots in a similar fashion to \citet{Reddy18} to create a \textit{Laggy Pilot} and a \textit{Noisy Pilot}. The laggy pilot is forced to repeat its previous action with probability $0.8$. The noisy pilot takes a uniform random action with probability $0.5$ and otherwise takes an action drawn from the optimal policy.

\subsection{Effects of Diverse Pilots in \lunarlander Domain}

We first test our method with a variety of simulated agents on the \lunarlander domain. Our goal is to understand the extent to which our copilots can assist different pilots as well as the effect that different surrogate human models have on training. To this end, we train separate copilots with each of the three surrogate human policies: a laggy pilot, a noisy pilot, and an imitation learning pilot. We identify copilots according to the pilot with which they are trained (e.g., \textit{laggy copilot}). We then compare the performance of the laggy, noisy, and imitation pilots when operating with each copilot as well as with no assistance. We evaluate each pilot-copilot team for $1000$ episodes. Detailed results can be found in Table~\ref{tab:lunar_sim}.

Across all pilot and copilot pairs, pilots perform better in terms of all three metrics when operating with the assistance of our copilot than they do without assistance. Perhaps unsurprisingly, we find that the laggy and imitation pilots perform best when paired with the associated copilot. However, we found that the noisy copilot improved the performance of the laggy pilot significantly more than the laggy copilot. We hypothesize that the noisy pilot provides greater diversity during training, which leads to a superior copilot. Interestingly, \citet[Table 1]{Reddy18} observed a similar phenomenon for the crash rate of their similarly defined laggy and noisy copilots. Additionally, we find that only the imitation copilot is able to prevent the imitation pilot from crashing. We explored training a copilot on a mixture of all three pilots, but initial experiments showed little improvement. We found slight performance gains for the laggy and noisy pilots, but a significant increase in the crash rate of the imitation pilot.

The above results empirically demonstrate that the pilot used during training has a large impact on the effectiveness of the copilot, which may pose challenges for our method on more complex tasks. There is a need for a training pilot that is ``similar enough'' to pilots seen at test time, and that simple human models like the laggy and noisy pilots do not effectively capture ``human-like'' behavior. Given that humans often adapt their behavior to robotic assistants, creating good human models can be a challenge. Future work might focus on how to make the copilot robust to any pilot it might encounter.

\subsection{Effects of Diverse Pilots in \dronereacher Domain}

In an effort to understand the effectiveness of our shared autonomy framework on more complex tasks, we train and evaluate different assistants in the \dronereacher environment. Again, our goal is to see how well our copilots can assist different pilots and what effect the surrogate human model has on training. We evaluate each pilot-copilot team for $1000$ episodes. Detailed results can be found in Table \ref{tab:drone_sim}.

Again, we find that for most pilot and copilot pairs, pilots perform better when operating with the assistance of a copilot than they do without assistance. However, we find that the noisy and imitation copilots are only marginally able to improve the performance of the laggy pilot. Additionally, the laggy copilot helps by providing a strong stabilizing force, preventing crashes at the expense of successes. However, even with this conservative approach, the laggy copilot is not able to completely prevent the imitation pilot from crashing. Only the imitation copilot is able to fully stabilize the imitation pilot. This matches the result from the \lunarlander experiments, highlighting the need for selecting a training pilot that is sufficiently similar to pilots seen at test time.

\subsection{Human Evaluation with \lunarlander}

\begin{figure}[!t]
    \centering
    \subfigure[\lunarlander]{\includegraphics[width=0.9\linewidth]{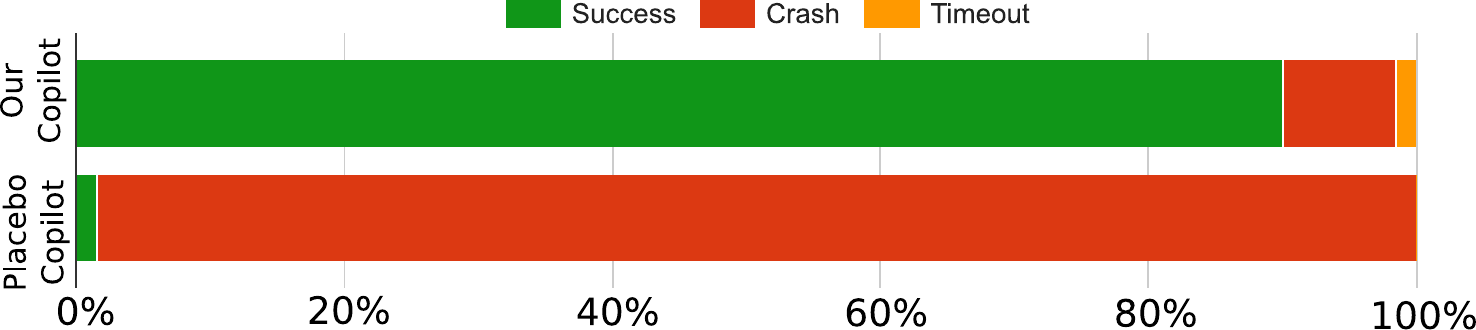}\label{fig:lunar_lander_success_crash}}\\
    \subfigure[\lunarreacher]{\includegraphics[width=0.9\linewidth]{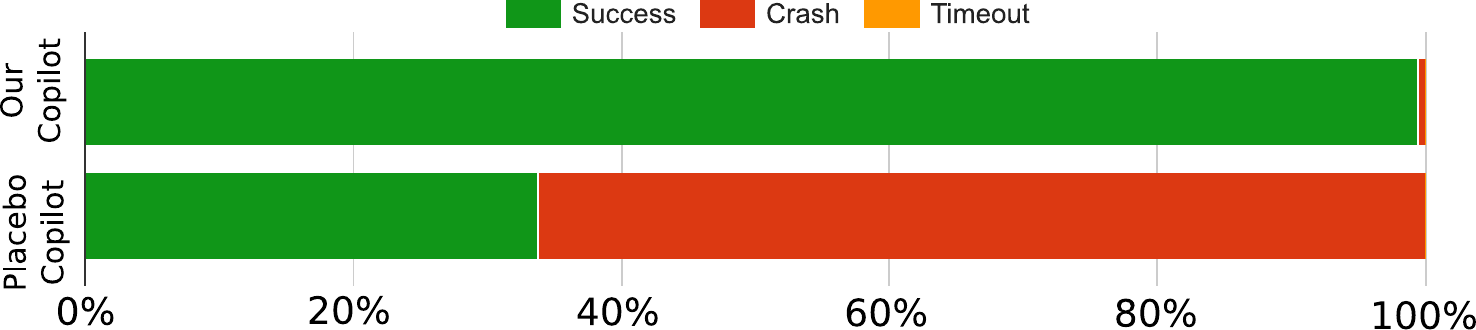}\label{fig:lunar_reacher_success_crash}}
    \subfigure[\dronereacher]{\includegraphics[width=0.9\linewidth]{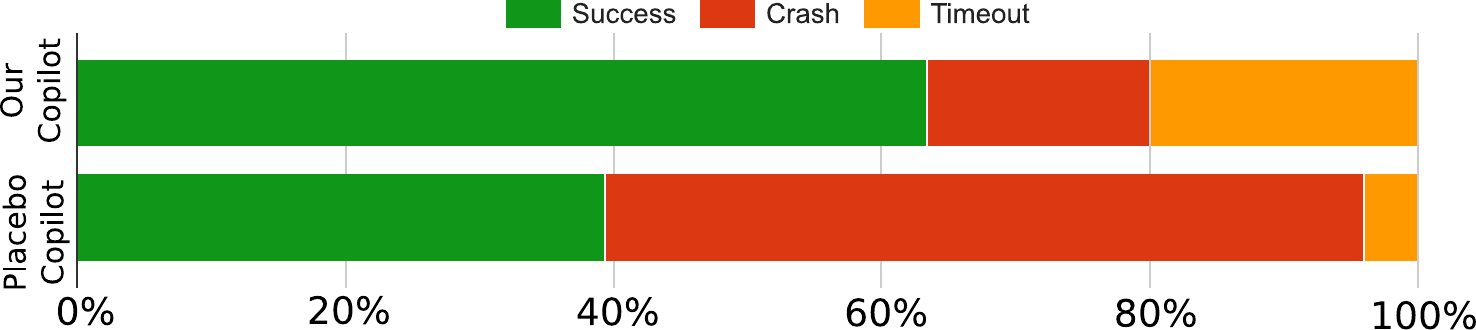}\label{fig:drone_reacher_success_crash}}\\
    \caption{A comparison of the performance of human pilots with and without the assistance of our shared autonomy agent in the \lunarlander, \lunarreacher, and \dronereacher domains. In each case, we see a significant increase in the success rate and a corresponding decrease in the crash rate when piloting the vehicle with the assistance of our agent.} \label{fig:human_success_crash}
\end{figure}
We perform a series of experiments to evaluate the effectiveness of our shared autonomy agent when used to assist human pilots. For the shared autonomy agent, we use the \lunarlander copilot that was trained with the imitation learning pilot. We did not fine-tune or otherwise specialize the copilot to the participants in any way. In addition to \lunarlander,
we additionally asked users to perform the \lunarreacher task, where the goal is to fly the spaceship to a randomly chosen floating position. This task requires different strategies that were not observed during training, i.e., flying to a top corner of the screen. Our copilot has no knowledge of which task the user is performing.

We recruited $16$ participants (all male, average age of $26$) to perform the evaluation. We manipulate two variables: the task (\lunarlander or \lunarreacher) and the copilot.
We ask users to interact with one of two assistants in each episode, one being our learned copilot and the other being a placebo copilot that does not modify their actions (i.e., $a_r \equiv 0$). The identity of the copilot is not known to user. To remove the confounding effect of the participants improving as they play, we randomize both the order of the tasks as well as the order of the copilots. Before each task, we allow the participant to practice without a copilot for $10$ episodes, with the option to practice for another $10$ episodes. We then allow the participant to practice with a copilot in the same way. Finally, we test the performance of the pilot-copilot pair over $20$ episodes. We repeat the process for each copilot and task.

We measure the performance of the copilot both quantitatively and qualitatively. Quantitatively, we compare the success and crash rates of each pilot-copilot team. Qualitatively, we ask participants to rank their agreement with five statements about each assistant as well as five statements comparing the two assistants. Absolute statements declared that a copilot was ``helpful,'' ``responsive,'' ``trustworthy,'' ``collaborated well,'' and ``performed consistently.'' Relative statements declared that one copilot ``made the task easier,'' ``was easier to learn,'' ``was more helpful,'' ``was preferred over,'' and ``was trusted more'' than the other copilot. With all statements, the copilots were identified only as ``copilot A'' and ``copilot B''. Participants reported their agreement on a five-point Likert scale.

\begin{figure}[!t]
    \centering
    \subfigure[\lunarlander]{\includegraphics[width=0.9\linewidth]{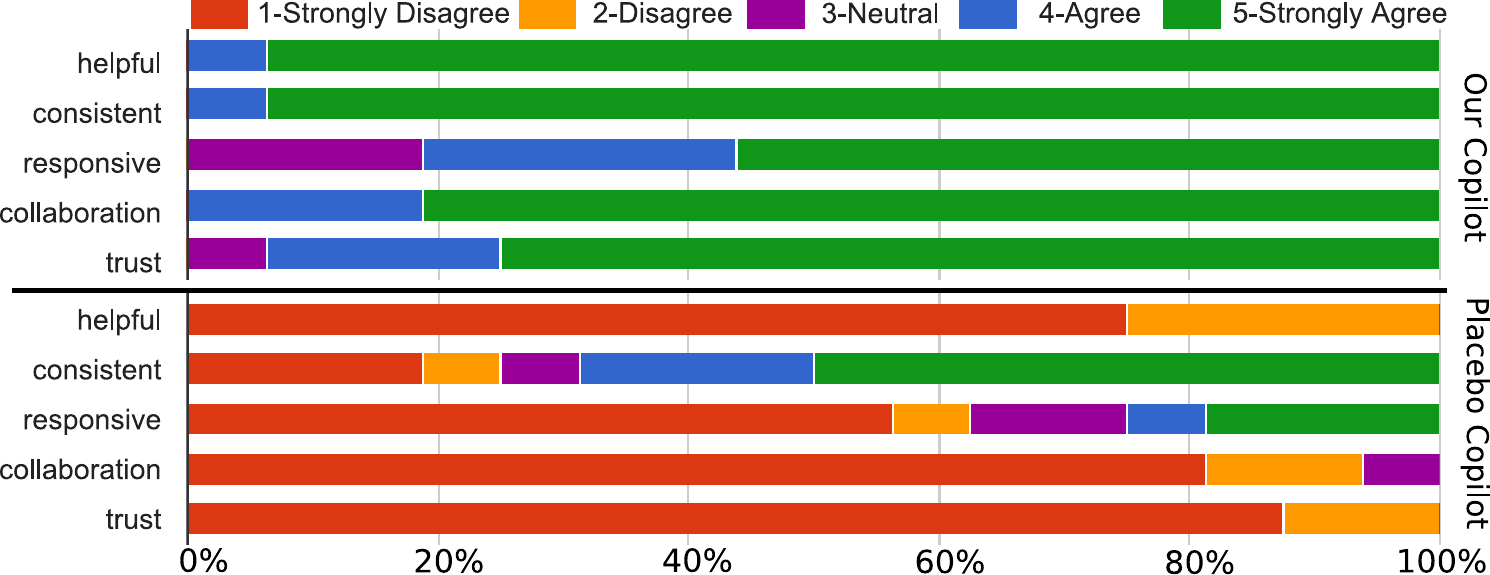}\label{fig:lunar_lander_qualitative}}\\
    \subfigure[\lunarreacher]{\includegraphics[width=0.9\linewidth]{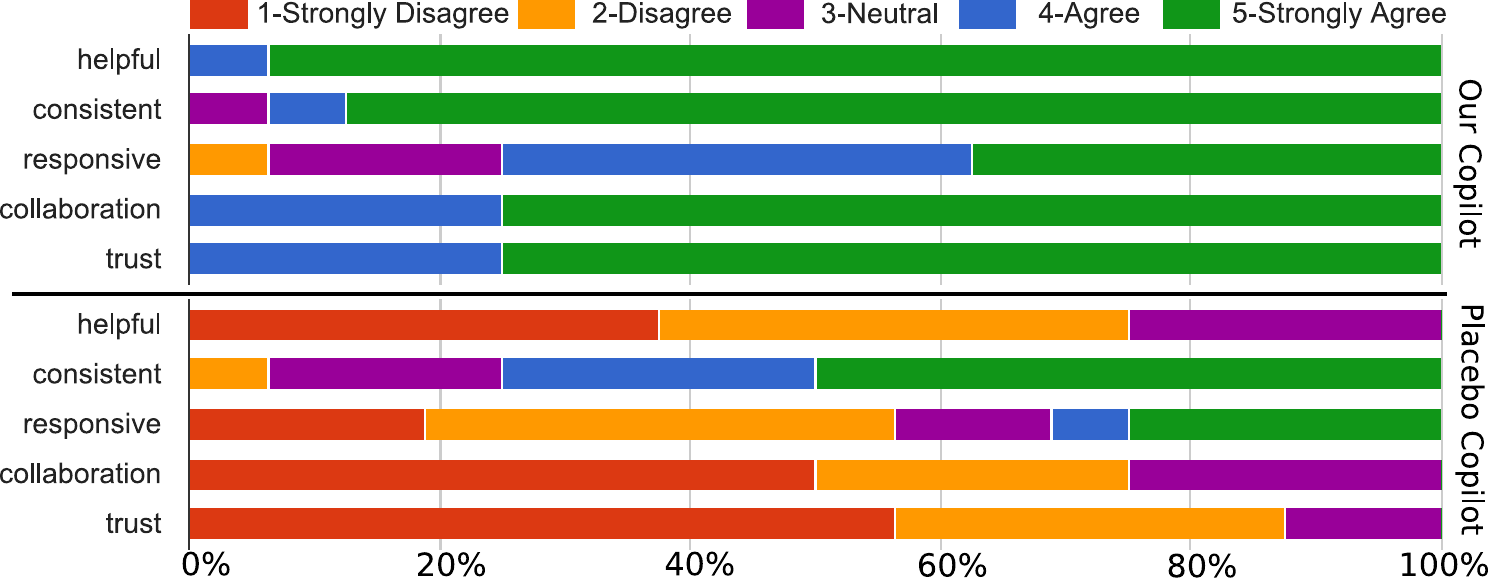}\label{fig:lunar_reacher_qualitative}}\\
    \subfigure[\dronereacher]{\includegraphics[width=0.9\linewidth]{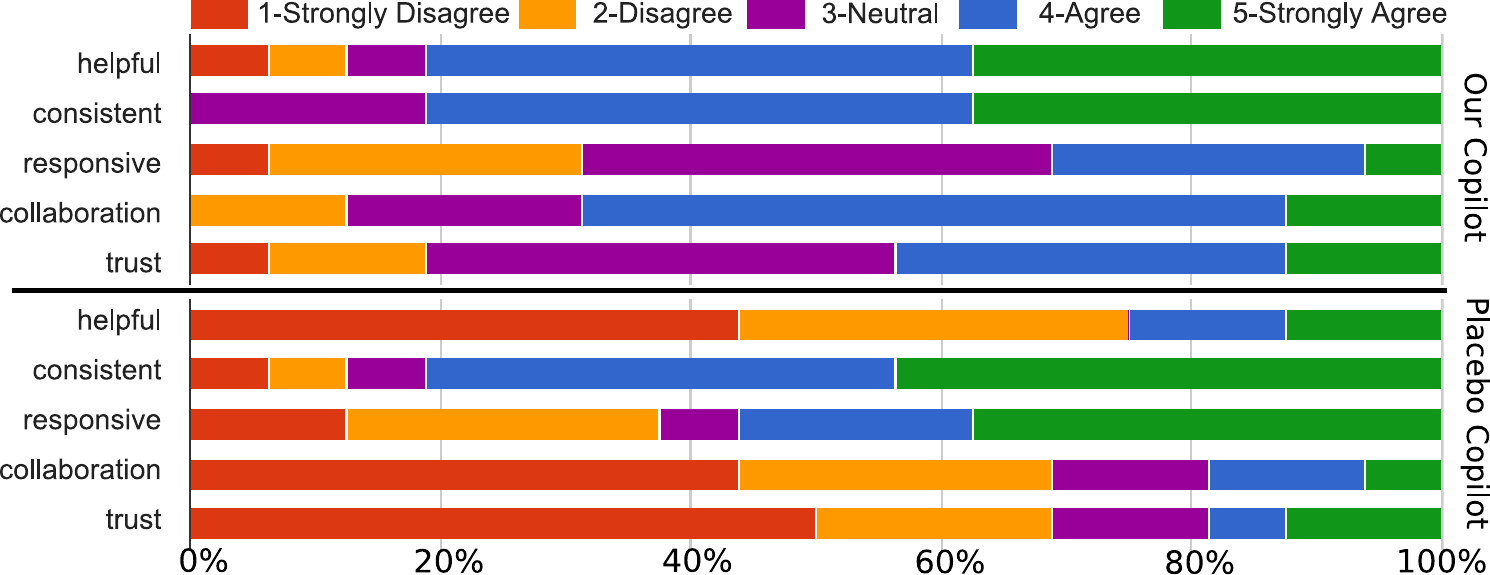}\label{fig:drone_reacher_qualitative}}
    \caption{Participants' response statistics for the human evaluation experiments for \subref{fig:lunar_lander_qualitative} \lunarlander \subref{fig:lunar_reacher_qualitative} \lunarreacher and \subref{fig:drone_reacher_qualitative} \dronereacher domains.}
    \label{fig:human_qualitative}
\end{figure}
Figure~\ref{fig:lunar_lander_success_crash} visualizes the performance statistics for the \lunarlander task. We find that our copilot significantly increases the success rate of the participants while also decreasing the crash rate compared to the no-assistance setting. We perform a Welch $t$-test for the success rate of each participant and find a maximum p-value across participants of $p = 1.02 \times 10^{-5}$. Therefore, we can easily reject the null hypothesis that our copilot has no effect on performance.
Qualitatively, as depicted in Figure~\ref{fig:lunar_lander_qualitative}, participants scored our copilot higher than the placebo in terms of helpfulness ($p = 2 \times 10^{-23}$), consistency ($p = 0.006$), responsiveness ($p = 7 \times 10^{-5}$), trustworthiness ($p = 3 \times 10^{-19}$), and sense of collaboration ($p = 5 \times 10^{-19}$). Interestingly, participants rated our copilot higher on consistency even though the placebo copilot very consistently did nothing. One explanation for this might be the participants' unfamiliarity with the environment dynamics.

We find similar results in the \lunarreacher environment (Fig.~\ref{fig:lunar_reacher_success_crash}), showing that our copilot is able to generalize across both tasks, despite having no explicit or implicit knowledge of the task. A Welch $t$-test for the success rate of each participant yields a maximum p-value across participants of $p = 0.004$. We find similar p-values for each qualitative question (Fig.~\ref{fig:drone_reacher_qualitative}) except for consistency, which is barely significant ($p = 0.03$).

Additionally, in both \lunarlander and \lunarreacher, participants agreed strongly with comparative statements in favor of our copilot over the placebo. In \lunarlander, every participant agreed (score of $4$ or $5$) that our copilot compared favorably to the placebo. In \lunarreacher, there was unanimous agreement for comparative statements of ease, helpfulness, and trustworthiness.

We also find that our method compares favorably to~\citet{Reddy18} in the \lunarlander environment. Unlike our framework, \cite{Reddy18} require access to the task reward function. Further, they fine-tune their copilot to each participant.
Additionally, our human pilots had an average success rate of $90\%$ while participants in~\cite[Fig. 3]{Reddy18} succeeded between $30\%$ and $80\%$ of the time. We emphasize that this is \emph{not} a strict comparison due to two main differences in the \lunarlander environment: we use a continuous action space while their action space is discrete, and they make the environment easier by modifying the vehicle's legs to make it more resistant to crashing on impact with the ground.

\subsection{Human Evaluation with \dronereacher}

We performed a similar human evaluation in the \dronereacher environment with $16$ participants ($14$ male and $2$ female, average age of $27$). We again test our copilot trained with the imitation learning surrogate human policy. We compare the participants' performance when operating the drone with our learned copilot and a placebo copilot (i.e., $a_r \equiv 0$). In order to
remove the confounding effect of the participants improving as they play, we randomize the order in which participants interact with each copilot. The structure of this evaluation is the same as that of \lunarlander and \lunarreacher: participants were evaluated with our copilot and a placebo copilot for $20$ episodes, after practicing alone and with both copilots. At the conclusion, participants were asked to rate their agreement with the same five absolute and five comparative statements.

As evident in Figure~\ref{fig:drone_reacher_success_crash}, we find that our copilot increases the success rate of the participants while significantly decreasing the crash rate compared to the placebo copilot. We perform a Welch $t$-test for the success and crash rates of each participant. We find that at a confidence level of $p = 0.05$, $7$ participants significantly improved their success rate and $11$ participants significantly improved their crash rate. Qualitatively, we found that participants scored our copilot higher than the placebo in terms of helpfulness ($ p = 0.0005 $), trustworthiness ($ p = 0.01 $), and sense of collaboration ($ p = 0.0004 $). Additionally, participants agreed with comparative statements in favor of our copilot over the placebo, giving an average score of $4$ or greater for all measures. One common observation was that our copilot often overrode user commands in favor of stabilizing the drone, leading to more timeouts at the task.

\section{Conclusion}
\label{sec:conclusion}

We have proposed a residual policy learning framework for shared autonomy with continuous action spaces. Our method does not assume knowledge of the environment dynamics, the space of possible goals, user policies, or require any task specific-knowledge. We only assume that there exists some reward function $\mathcal{R}_{\text{general}}$ that expresses goal-agnostic objectives common to a diversity of tasks. Our method learns an assistive agent optimized to minimize its interference while satisfying a predefined set of constraints on the goal-agnostic reward. This effectively endows the human with the responsibility for goal-specific actions, while allowing the agent to correct these actions when they would violate a goal-agnostic constraint.

We evaluated our method on three continuous control tasks: \lunarlander, \lunarreacher, and \dronereacher (a reaching task with a quadrotor drone simulator). We find that our learned agents greatly improve the performance of both simulated actors and human operators. Additionally, humans qualitatively rank our agent significantly higher than a placebo agent that does not provide assistance.

One limitation of our work is the need for a human model during training. This can be difficult to engineer given that humans adapt their behavior to the assistant. A possible direction for future work is to make the learned agent less sensitive to the human model used during training and therefore make it more robust to different users and tasks. Additionally, the model-free nature of our framework brings with it greater sample complexity. While similar model-free shared autonomy methods have been shown to be applicable to real hardware~\cite{Reddy18}, future work will explore recent results in sim-to-real transfer for complex control tasks~\cite{peng18,tan18} as a means of extending our method to physical robots. Further, while we tested our algorithm in environments that required stabilizing flight, our algorithm could be applied to other applications such as assisted driving or manipulation.

\ifarxiv
\section*{Acknowledgments}

This work was supported in part by the National Science Foundation under Grant No.\ 1830660.

\fi

\bibliographystyle{plainnat}
{\small
\bibliography{references}
}

\end{document}